\documentclass[accepted]{article}

\usepackage{microtype}
\usepackage{graphicx}
\usepackage{subfigure}
\usepackage{booktabs} 

\usepackage{hyperref}

\usepackage{icml2020}

\icmltitlerunning{Towards Robust Classification with Deep Generative Forests}

\usepackage{graphicx}
\usepackage{subfigure}
\usepackage{mathtools}
\usepackage{bm}
\usepackage{bbm}
\usepackage{amsthm}
\usepackage{nicefrac}
\usepackage{amsmath}
\usepackage{amsfonts}
\usepackage{amssymb}
\usepackage{pgf}
\usepackage{tikz}
\usetikzlibrary{arrows, calc}
\usepackage{scalerel}

\DeclareMathOperator*{\ch}{ch}

\newcommand{\cbar}{\,|\,}
\newcommand{\X}{\mathbf{X}}
\newcommand{\x}{\mathbf{x}}

\newcommand{\data}{\mathcal{D}}

\newcommand{\indf}[1]{\ensuremath{{\mathbbm{1}(#1)}}}
\newcommand{\indfu}[1]{\ensuremath{{\mathbbm{1}_{#1}}}}
\newcommand{\node}{v}
\newcommand{\nodec}{u}

\newcommand{\w}{w}

\newcommand{\ntrees}{\ensuremath{n_t}}

\newcommand{\GeDT}{GeDT}
\newcommand{\GeF}{GeF}
\newcommand{\Gp}{GeF$^+$}

\begin{document}

\twocolumn[
\icmltitle{Towards Robust Classification with Deep Generative Forests}



\icmlsetsymbol{equal}{*}

\begin{icmlauthorlist}
\icmlauthor{Alvaro H. C. Correia}{tue}
\icmlauthor{Robert Peharz}{tue}
\icmlauthor{Cassio de Campos}{tue}
\end{icmlauthorlist}

\icmlaffiliation{tue}{Eindhoven University of Technology}

\icmlcorrespondingauthor{Alvaro H. C. Correia}{a.h.chaim.correia@tue.nl}

\icmlkeywords{Reliable Machine Learning, Outlier Detection, Robustness, Uncertainty, ICML}

\vskip 0.3in
]



\printAffiliationsAndNotice{} 

\begin{abstract}
Decision Trees and Random Forests are among the most widely used machine learning models, and often achieve state-of-the-art performance in tabular, domain-agnostic datasets. Nonetheless, being primarily discriminative models they lack principled methods to manipulate the uncertainty of predictions. 
In this paper, we exploit Generative Forests (GeFs), a recent class of deep probabilistic models that addresses these issues by extending Random Forests to generative models representing the full joint distribution over the feature space.
We demonstrate that GeFs are uncertainty-aware classifiers, capable of measuring the robustness of each prediction as well as detecting out-of-distribution samples.
\end{abstract}

\section{Introduction}
Decision Trees (DTs) and Random Forests (RFs) are arguably the most popular non-linear machine learning models of today.
In Kaggle's 2019 report on the \emph{State of Data Science and Machine Learning} \cite{Kaggle2019}, DTs and RFs appear as second most widely used techniques, right after linear and logistic regressions.
Moreover, decision trees are often considered interpretable \cite{Freitas2014} and hence have enjoyed a surge in popularity with the increasing interest in explainable artificial intelligence.
Nonetheless, efforts towards uncertainty-aware and reliable tree-based models are still comparatively scarce.

In this paper, we demonstrate that some of these shortcomings are addressed by Generative Forests (\GeF{}s) \cite{Correia2020}, a class of deep probabilistic models that subsumes Random Forests.
In particular, we show in a number of classification tasks that \GeF{}s enable new principled methods to i) estimate the uncertainty of \emph{each} of the model's predictions and ii) monitor the input distribution to detect out-of-domain samples or distribution shifts.

\section{Generative Forests}
Before discussing the main ideas of the paper, we introduce Generative Forests and the required notation.
As we focus on classification tasks, we denote the set of explanatory variables as $\X = \{X_1, X_2, \ldots, X_m\}$ and the target variable as $Y$. 
As usual, we write realisations of random variables (or collections thereof) in lowercase; for example, $\X=\x$ or $Y=y.$
We assume the pair $(\X,Y)$ is drawn from a fixed joint distribution $\mathbb{P}^*(\X, Y)$ with density $p^*(\X, Y)$ and that, while the true distribution $\mathbb{P}^*$ is unknown, we have a dataset $\data_n = \{(\x_1,y_1), \ldots, (\x_n,y_n)\}$ of $n$ i.i.d.~samples from $\mathbb{P}^*$.

Generative Forests are in fact a class of \emph{Probabilistic Circuits} (PCs) \cite{VanDenBroeck2019} satisfying smoothness and decomposability~\cite{Peharz2015}. PCs are a family of deep density representations facilitating many exact and efficient inference routines \cite{Darwiche2003, VanDenBroeck2019}.
In short, they are computational graphs with three types of nodes:
i) \emph{distribution nodes}, ii) \emph{sum nodes} and iii) \emph{product nodes}.
Distribution nodes compute a probability density (by an adequate choice of the underlying measure, this also subsumes probability mass functions) over some subset $\X' \subseteq \X$, that is, a normalised function mapping the state space of $\X'$ to the non-negative real numbers.
Sum nodes compute convex combinations over their children: if $\node$ is a sum node and $\ch(\node)$ its children, then $\node$ computes $\node(\x) = \sum_{\nodec \in \ch(\node)} \w_{\node,\nodec} \nodec(\x)$, where $\w_{\node,\nodec} \geq 0$ and $\sum_{\nodec \in \ch(\node)} \w_{\node,\nodec} = 1.$
Product nodes compute the product over their children; if $\node$ is a product node, then $\node(\x) = \prod_{\nodec \in \ch(\node)} \nodec(\x_u)$, with the collection $\{\x_u\}_{u\in\ch(\node)}$ a partition (non-overlapping projections) of $\x$.
Finally, a (smooth and decomposable) PC represents the density over all variables (here $p(\X, Y)$) computed by its root node.

Generative Forests are best understood by relating individual decision trees to Probabilistic Circuits.
For any given DT, we can construct a corresponding PC---a \emph{Generative Decision Tree} (\GeDT{})---representing a full joint density $p(\X,Y)$.
In a nutshell, each decision node is converted into a sum node and each leaf into a density with support restricted to the leaf's cell.
The training samples can be figured to be routed from the root node to the leaves, following the decisions at each decision/sum node.
The sum weights are given by the fraction of samples which are routed from the sum node to each of its children.
The leaf densities are learned on the data which arrives at the respective leaves.

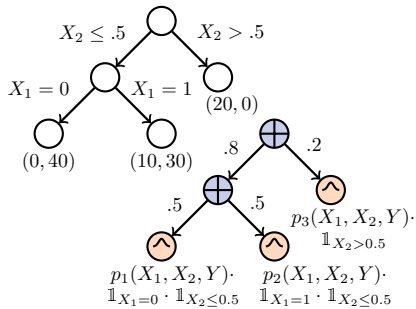
\begin{figure}[t!]
    \begin{center}

\definecolor{blu}{RGB}{199, 206, 234}
\definecolor{gre}{RGB}{181, 234, 215}
\definecolor{re}{RGB}{255, 154, 162}
\definecolor{ore}{RGB}{255, 218, 193}
\definecolor{lgr}{RGB}{226, 240, 203}
\definecolor{mel}{RGB}{255, 183, 178}

\begin{tikzpicture}[thick,scale=0.75, every node/.style={transform shape},
    roundnode/.style={minimum size=7.mm,
                      inner sep=0}, 
    cross/.style={minimum size=5mm, fill=lgr,
        path picture={
            \draw[black] (path picture bounding box.south east) -- (path picture bounding box.north west) (path picture bounding box.south west) -- (path picture bounding box.north east);
        }
    },
    sum/.style={minimum size=5mm, fill=blu,
        path picture={
            \draw[black] (path picture bounding box.south) -- (path picture bounding box.north) (path picture bounding box.west) -- (path picture bounding box.east);
        }
    },
    gauss/.style={minimum size=5mm, fill=ore,
        path picture={
            \draw[black] plot[domain=-.15:.15] ({\x},{exp(-200*\x*\x -2.)});
        }
    },
    dt/.style={minimum size=5mm, fill=white
    },
    line/.style={
      draw,thick,
      -latex',
      shorten >=2pt
    },
    cloud/.style={
      draw=red,
      thick,
      ellipse,
      fill=red!20,
      minimum height=1em
    }
]
    
    \node[draw, circle, dt] (D1) at (4, 3) { };
    \node[draw, circle, dt] (D2) at (3, 2) { };
    \node[draw, circle, dt] (D3) at (5, 2) { };
    \node[draw, circle, dt] (D4) at (2, 1) { };
    \node[draw, circle, dt] (D5) at (4, 1) { };
    
    \node (ind) at (5.225, 2.8) {$X_2>.5$};
    \node (ind) at (4.0, 1.775) {$X_1=1$};
    \node (ind) at (5.25, 1.5) {$(20,0)$};
    \node (ind) at (4.0, 0.5) {$(10,30)$};
    \node (ind) at (2.0, 0.5) {$(0,40)$};

    \draw[line width=0.3mm, <-, auto] (D2.45) to node[black]{$X_2\leq.5$} (D1.225);
    \draw[line width=0.3mm, <-] (D3.135) -- (D1.315);
    \draw[line width=0.3mm, <-, auto] (D4.45) to node[black]{$X_1=0$} (D2.225);
    \draw[line width=0.3mm, <-] (D5.135) --  (D2.315);

    \node[draw, circle, sum] (Dz1) at (4+2, 3-2) { };
    \node[draw, circle, sum] (Dz2) at (3+2, 2-2) { };
    \node[draw, circle, gauss] (Dz3) at (5+2, 2-2) { };
    \node[draw, circle, gauss] (Dz4) at (2+2, 1-2) { };
    \node[draw, circle, gauss] (Dz5) at (4+2, 1-2) { };
    
    \node (ind) at (4.7+2, 2.8-2) {$.2$};
    \node (ind) at (3.7+2, 1.775-2) {$.5$};
    \node (ind) at (5.4+2, 1.5-2) {$p_3(X_1,X_2,Y)\cdot$};
    \node (ind) at (5.4+2, 1.1-2) {$\indfu{X_2 > 0.5}$};
    \node (ind) at (4.4+2.5, 0.5-2) {$p_2(X_1,X_2,Y)\cdot$};
    \node (ind) at (4.4+2.5, 0.1-2) {$\indfu{X_1=1}\cdot \indfu{X_2\leq 0.5}$};
    \node (ind) at (1.7+2.5, 0.5-2) {$p_1(X_1,X_2,Y)\cdot$};
    \node (ind) at (1.7+2.5, 0.1-2) {$\indfu{X_1=0}\cdot \indfu{X_2\leq 0.5}$};

    \draw[line width=0.3mm, <-, auto] (Dz2.45) to node[black]{$.8$} (Dz1.225);
    \draw[line width=0.3mm, <-] (Dz3.135) -- (Dz1.315);
    \draw[line width=0.3mm, <-, auto] (Dz4.45) to node[black]{$.5$} (Dz2.225);
    \draw[line width=0.3mm, <-] (Dz5.135) --  (Dz2.315);

\end{tikzpicture}
\end{center}
    \vspace{-.25cm}
    \caption{Illustration of a DT and its corresponding PC.}
    \label{fig:dt-spn}
\end{figure}

Note that \GeDT{}s are proper PCs over $(\X,Y)$, albeit rather simple ones: they are tree-shaped and contain only sum nodes.
Nonetheless, \GeDT{}s are in fact a class of models, as we are free to fit arbitrarily complex functions at the leaves; say graphical models, again PCs or even advanced density estimators such as a VAEs \cite{Kingma2014} or Flows \cite{Rezende2015}.
In this work, however, we focus on arguably the simplest density estimator, and model the density at the leaves as $p(\X, Y)=p(X_1)\ldots p(X_m)p(Y)$, with continuous and categorical variables represented by univariate normal and multinomial distributions, respectively.
We show these \emph{fully-factorised} leaves are already sufficient to equip standard RFs with effective and principled ways to detect outliers and estimate the robustness of each prediction.

The main semantic difference between DTs and \GeDT{}s is that a DT represents a classifier, that is, a conditional distribution $f(\x)$, while the corresponding \GeDT{} encodes a full joint distribution $p(\X, Y)$---the latter naturally lends itself to classification via the conditional distribution $p(Y \cbar \x) \propto p(\x, Y)$.
Note that, in theory, $p(Y \cbar \x)$ might differ substantially from $f(\x)$, as every feature might influence classification in a \GeDT{}, even if it never appears in any decision node of the DT.
Still, it is easy to see that if the distribution at the leaves satisfy $p(\X, Y)=p(\X)p(Y)$, then a \GeDT{} defines the same prediction function as the original DT.

Generative Forests are ensembles of \GeDT{}s and can also be made equivalent to the original RF by an appropriate choice of density model at the leaves. However, instead of ensuring ``backwards compatibility'', in this paper we are interested in exploiting the generative properties of \GeF{}s.
To that end, we extend \GeF{}s to model a single joint by considering a uniform mixture of \GeDT{}s (using a sum node over the trees). That is, instead of averaging over the conditional distributions of each of the \ntrees{} trees, we define a single model that represents the joint $p(\X,Y)=\ntrees^{-1}\sum_{j=1}^{\ntrees} p_j(\X,Y)$, where each $p_j$ comes from a \GeDT. Since this model is essentially a mixture of the different trees, we call it \Gp{}.

\section{Outlier Detection}
Most of machine learning theory relies on the assumption that training and test data are sampled from the same distribution. This is a reasonable assumption---there would be no hope for learning otherwise---but is often violated in practice, as real-world data is constantly evolving. 
Reliable machine learning models should then be able to identify such violations to either suspend judgement and fail gracefully or signal the need for further data gathering and retraining.

\begin{figure}[h!]
    \begin{center}
        \scalebox{0.9}{\input{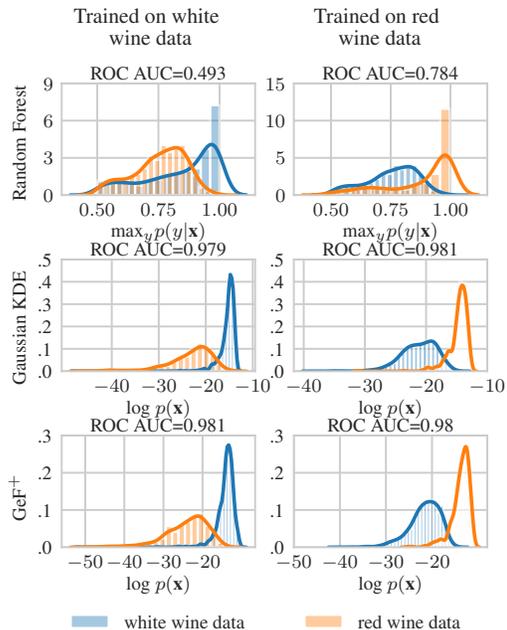}}
    \end{center}
    \vspace{-.5cm}
    \caption{Normalised histograms of $\log p(\x)$ (KDE and \Gp) and $\max_y p(y|\x)$ (RF) of samples from red and white wine data.}
    \label{fig:fx_wine}
\end{figure}

Generative models offer a natural and principled way to detect outliers or distribution shifts. As they innately fit the joint distribution of the training data, they are capable of estimating the likelihood of every new sample, flagging unlikely ones as potential anomalies.
In a \Gp{} this is done by monitoring the marginal $p(\X)$, which comes at no extra cost; classification is performed over the joint $p(Y, \X)$ and computing $p(\x)$ only requires summing over all possible classes, $p(\x) = \sum_y p(y, \x)$.

We illustrate outlier detection in \Gp{}s using the wine quality dataset \cite{Cortez2009} (where the class is a scale of quality of wine) with a variant of transfer testing \cite{Bradshaw2017}. We learn two different \Gp{}s, each with only one type of wine (red or white), and compute the log-density of unseen data (70/30 train-test split) for the two wine types. 
As we see in the histograms of Figure~\ref{fig:fx_wine}, the marginal distribution over the joints does provide a strong signal to identify out-of-domain instances. 
We compare \Gp{}s to a Gaussian Kernel Density Estimator (KDE) and to a common baseline for deep models \cite{hendrycks2016}, whereby the probability of the predicted class, $\max_y p(y|\x)$, is used as a signal to detect outliers.
We see from the histograms and the ROC (receiver operating characteristic curve) scores, that our models largely outperform the baseline while being comparable to KDEs, even though the structure of a \Gp{} is learned in a discriminative manner.

We note that previous works have already proposed using Random Forests for outlier detection \cite{Liu2008}. 
However, these models are typically directly trained to identify anomalies and have that as their sole purpose, while \Gp{}s are unique in that, while being primarily classifiers (or regressors) they also effectively detect out-of-domain samples.

\section{Robust Classification}
Outlier detection is related to the concept of \emph{vagueness} or \emph{epistemic uncertainty}, that is, the lack of sufficient statistical support for issuing a prediction. 
However, machine learning models are often confronted with cases where the data supports the thesis that a given instance is associated with more than a single class with high probability. That is commonly referred to as \emph{aleatory uncertainty}.
Effectively quantifying both types of uncertainty is indispensable in critical applications, where overconfident predictions may lead to catastrophic failures.
One common approach to estimate the model's confidence in classification tasks is to manipulate the reported probability $p(Y|\x)$ \cite{Guo2017,Liang2017}. 
Still, this is not only overly simplistic but also fails to distinguish the types of uncertainty.

\begin{figure}[h!]
    \begin{center}
        \scalebox{0.8}{\input{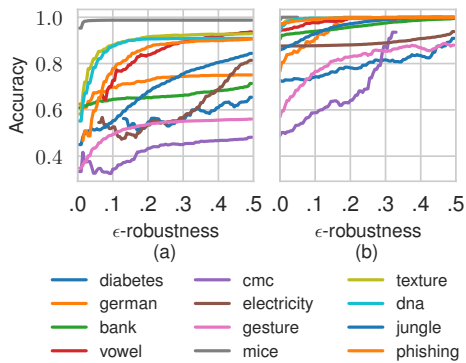}}
    \end{center}
    \vspace{-.75cm}
    \caption{Accuracy of predictions with $\epsilon$-robustness (a) below and (b) above different thresholds for 12 OpenML datasets. Some curves end abruptly because we only computed the accuracy when 30 or more data points were available for a given threshold.}
    \label{fig:acc_rob}
\end{figure}

\Gp{}s offer an arguably more principled approach rooted in the notion of \emph{robustness} \cite{Dietterich2017} as obtained with credal sum-product networks~\cite{Maua2017,Maua2018}.
In a nutshell, we evaluate how much we can perturb all parameters of the model without changing its prediction on a given instance. 
Formally, we quantify this perturbation with the concept of $\epsilon$-contamination for each of the sum nodes in a PC.
If $w$ is the vector of weights of a given sum node, then its $\epsilon$-contamination is given by the set
$$
\mathcal C_{w, \epsilon} = \{(1-\epsilon)w + \epsilon v: v_j \geq 0, \sum_j v_j = 1\}.
$$
This definition naturally leads to the idea of $\epsilon$-robustness: the largest $\epsilon$ for which all parameter configurations in $\mathcal C_{w, \epsilon}$ yield the same classification. We run such analysis for all of the nodes at once: let $\mathcal{C}_{\epsilon}$ represent the collection $\{\mathcal C_{w, \epsilon}\}$ for all sum nodes in the PC and let $\mathbf{w}$ be one possible choice of a $w$ in each of the nodes.\footnote{Multinomial leaf nodes can be contaminated in the very same manner as sum nodes, while normal leaf nodes are contaminated in their means while keeping variance fixed~\cite{dewit19a}.} We compute whether there is a label $y$ of the class such that
\begin{equation*}
    \forall y'\neq y: \max_{\mathbf{w}\in\mathcal{C}_{\epsilon}} \mathbb{E}_{\mathbf{w}}[\indf{Y=y'} - \indf{Y=y}|\mathbf{x}] < 0\, ,
    \label{eq:rob}
\end{equation*}
\noindent and if so, we declare $y$ robust for threshold $\epsilon$. 
The maximum $\epsilon$ such that this is true (which we find by binary search) is what we call the $\epsilon$-robustness of a given prediction ~\cite{Maua2018}.
Note that, since \Gp{}s have a tree structure and sum nodes with out-degree bounded by a constant, the time for computing $\epsilon$-robustness in \Gp{}s is linear in the input size \cite{Correia2019,Maua2018}. 

We experiment with $\epsilon$-robustness in a selection of 12 datasets from the OpenML-CC18 benchmark\footnote{\url{https://www.openml.org/s/99/data}} \cite{OpenML2013}. 
Once more, we use \Gp{}s composed of 30 trees with fully-factorised leaves and a 70/30 train-test split.
In Figure~\ref{fig:acc_rob}, we defined a number of robustness thresholds and, for each of them, we computed the accuracy of the models over instances for which $\epsilon$ was above and below the threshold. We clearly see there is a positive correlation; the higher the $\epsilon$-robustness of a prediction, the more likely it is to be correct. Obviously, the computation of robustness does not require knowing the true labels.

This concept of robustness has a clear interpretation. Given that for $\epsilon=0$ we have $\mathcal C_{w, \epsilon}=\{w\}$ and for $\epsilon=1$ we have the whole simplex, we can interpret the value of $\epsilon$ as the ``percentage'' of variation that we allow in the parameters for \emph{each prediction}. 
That is in contrast to typical uncertainty measures where individual uncertainty values are hard to interpret in isolation \cite{Mentch2016, Shaker2020}.

\section{Discussion and Further Experiments}

\begin{figure}[t!]
        \centering
        \begin{subfigure}
            \centering
            \includegraphics[width=.225\textwidth]{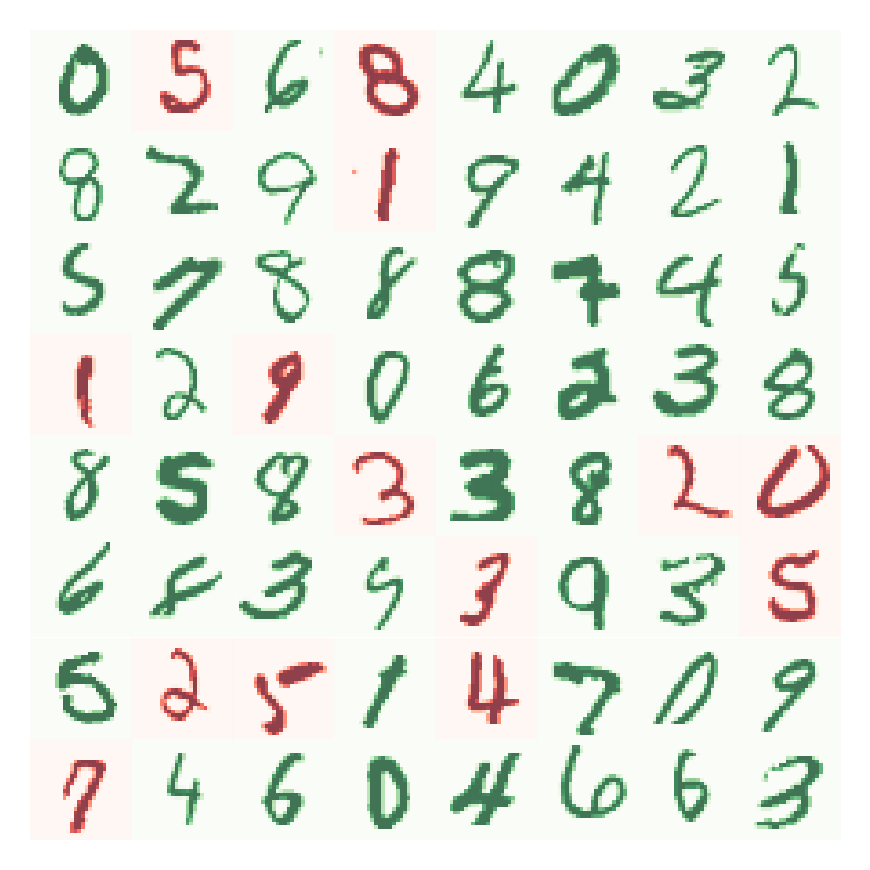}
        \end{subfigure}
        \begin{subfigure}
            \centering
            \includegraphics[width=.225\textwidth]{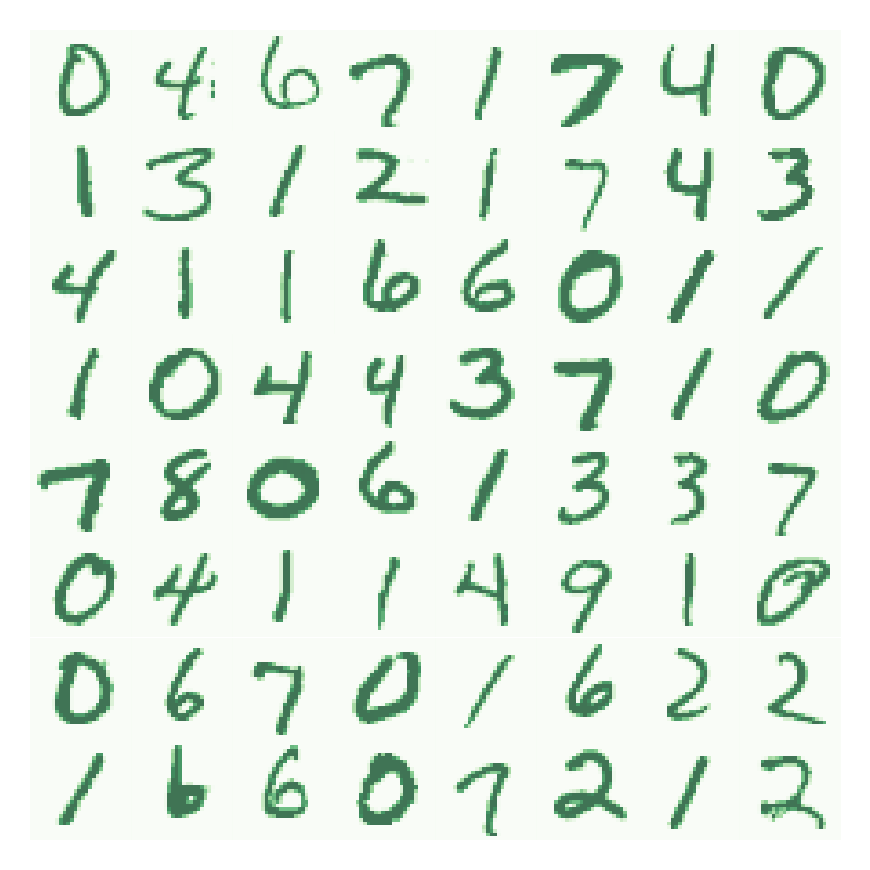}
        \end{subfigure}
        \hfill
        \vskip -10.pt
        \hfill
        \begin{subfigure}
            \centering
            \includegraphics[width=.225\textwidth]{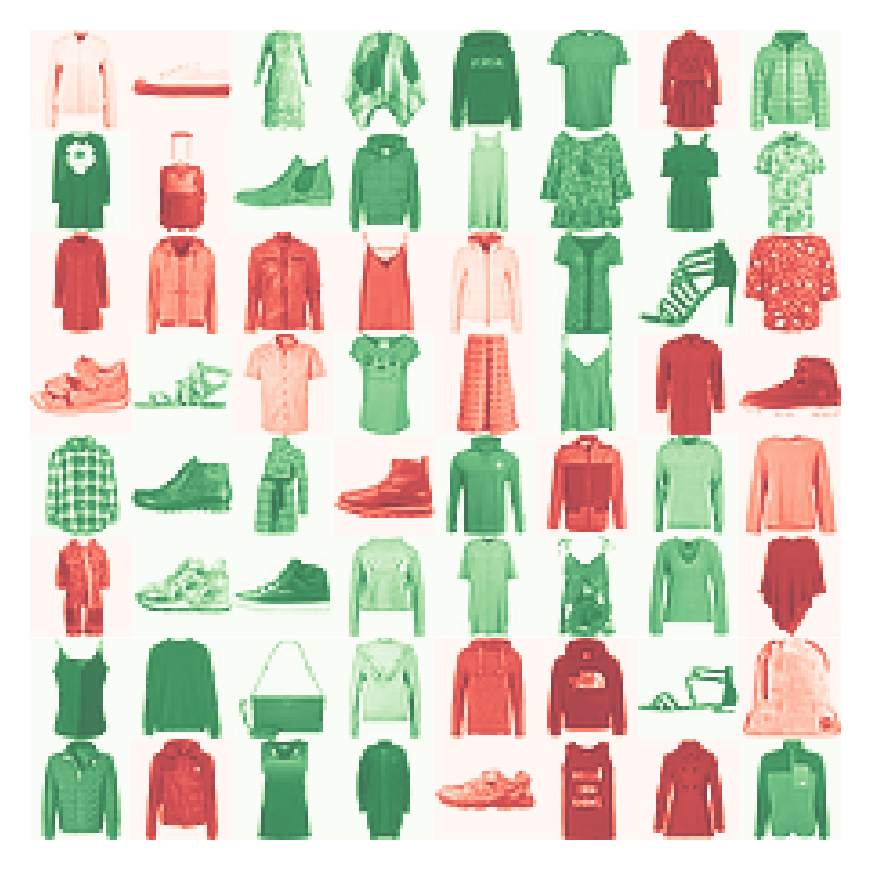}
        \end{subfigure}
        \begin{subfigure}
            \centering
            \includegraphics[width=.225\textwidth]{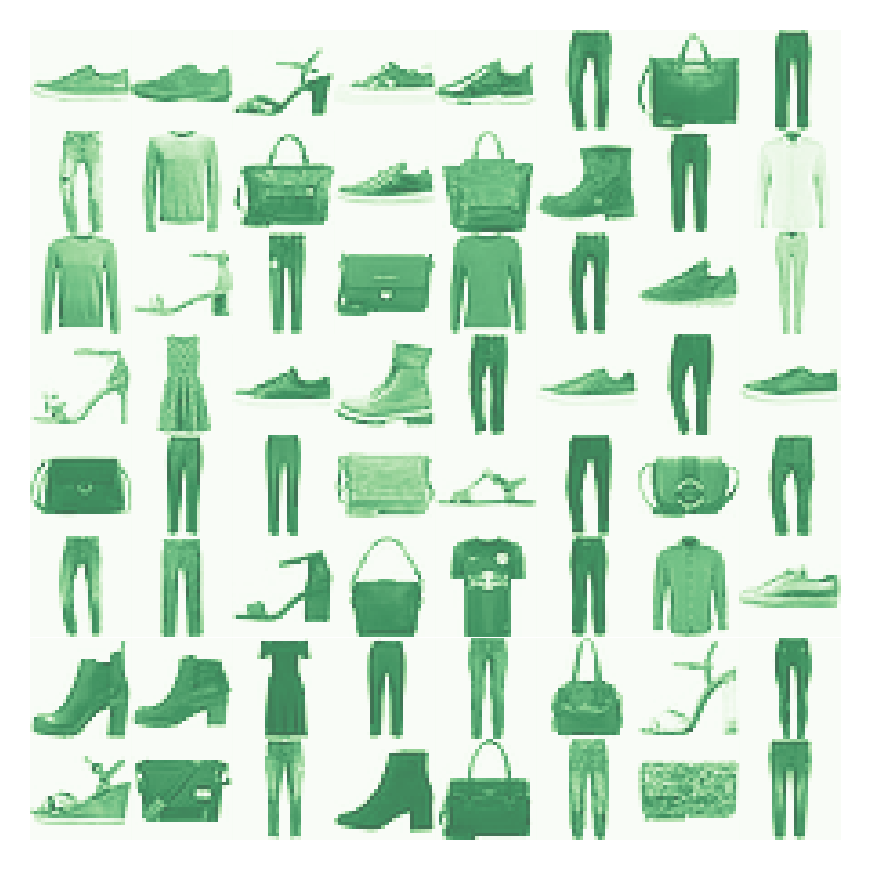}
        \end{subfigure}
        \hfill
        \vspace{-0.35cm}
        \caption{Samples from (Fashion-)Mnist datasets with lowest (left) and highest (right) $\epsilon$-robustness in the test set. Correctly and incorrectly classified examples are shown in green and red, respectively.}%
    \label{fig:mnist}%
\end{figure}

We also trained \Gp{}s on the Mnist \cite{Lecun2010} and Fashion-Mnist \cite{xiao2017} datasets to visually evaluate the samples with different $\epsilon$-robustness values. 
In Figure~\ref{fig:mnist}, we report test instances with lowest and highest $\epsilon$-robustness for each dataset. We see that samples for which the prediction is less robust are not only less likely to be correctly classified but also often contain irregular shapes and patterns, justifying the model's uncertainty.

\begin{figure}[t]
        \centering
        \begin{subfigure}
            \centering
            \includegraphics[width=.225\textwidth]{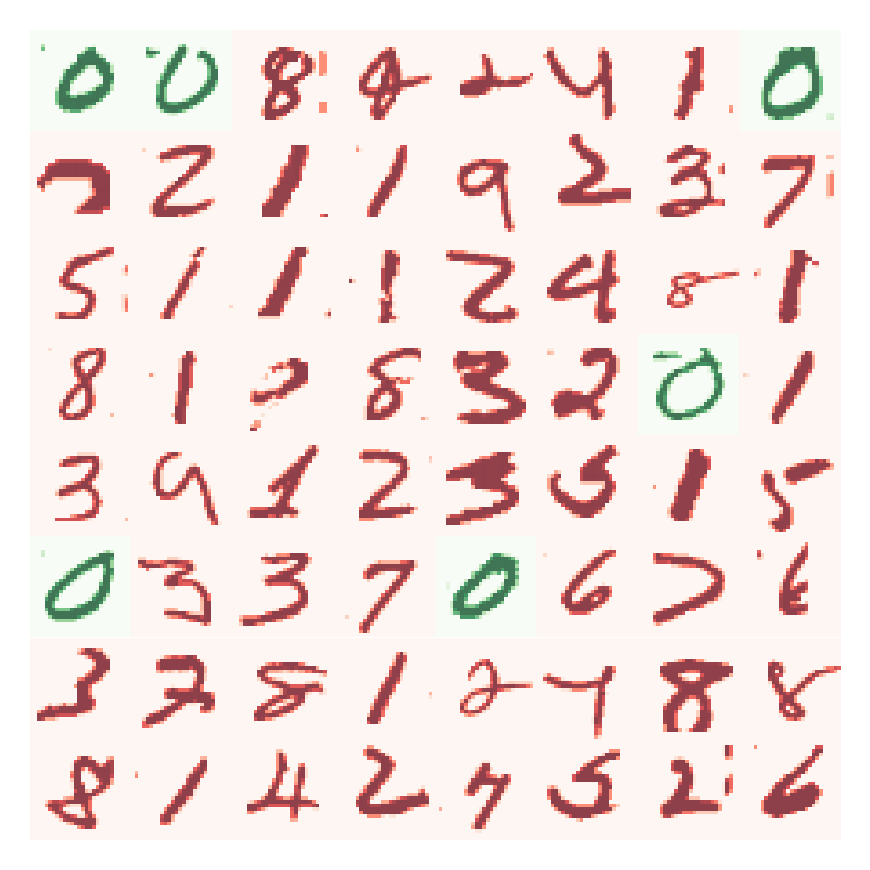}
        \end{subfigure}
        \begin{subfigure}
            \centering
            \includegraphics[width=.225\textwidth]{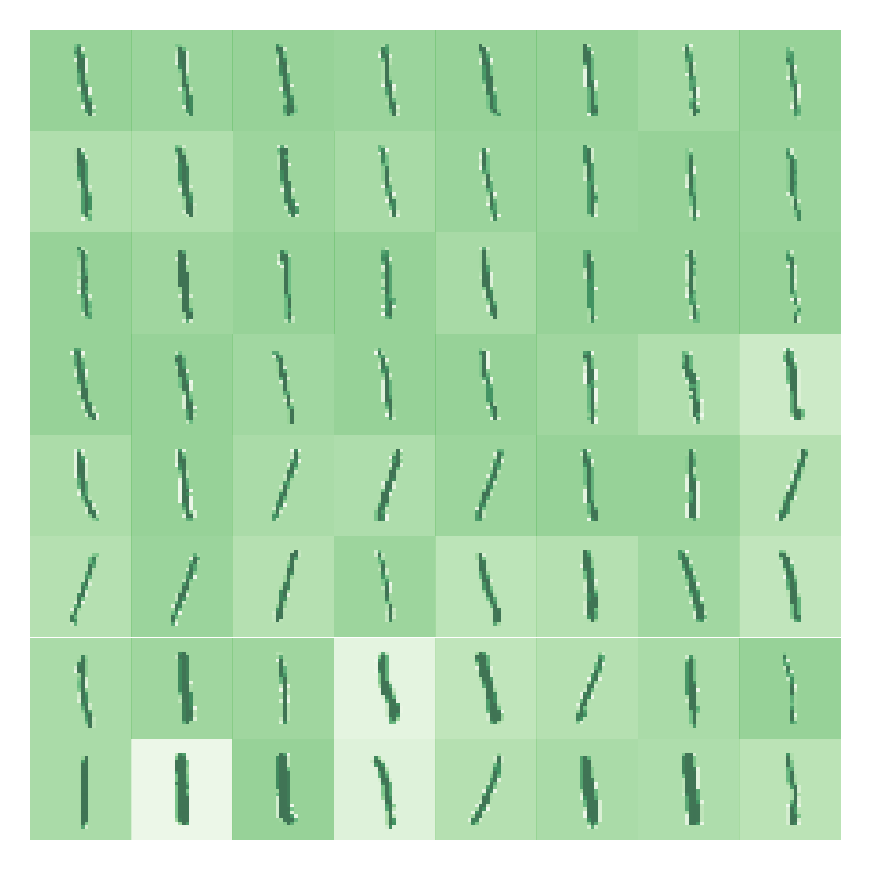}
        \end{subfigure}
        \hfill
        \vskip -10.pt
        \hfill
        \begin{subfigure}
            \centering
            \includegraphics[width=.225\textwidth]{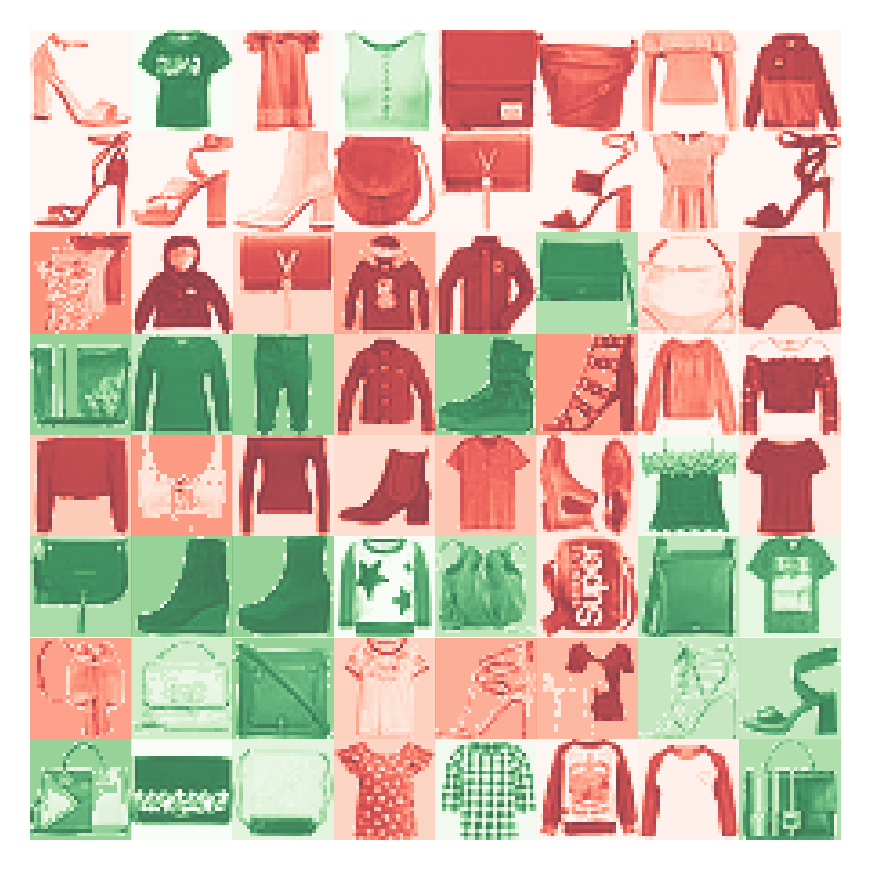}
        \end{subfigure}
        \begin{subfigure}
            \centering
            \includegraphics[width=.225\textwidth]{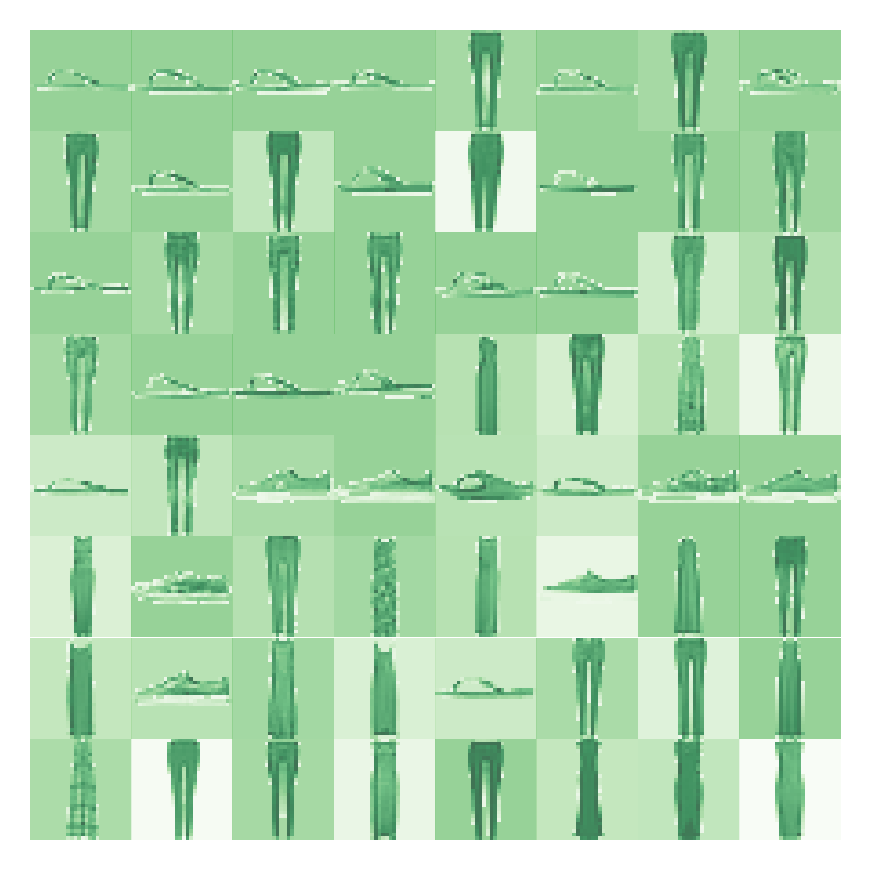}
        \end{subfigure}
        \hfill
        \vspace{-0.35cm}
        \caption{Samples from (Fashion-)Mnist datasets with lowest (left) and highest (right) $p(\x)$ in the test set. The background light is proportional to the $\epsilon$-robustness.}%
    \label{fig:log_mnist}%
\end{figure}

We emphasise outlier detection and robustness estimation are related but different notions, and \Gp effectively distinguishes them.
Figure~\ref{fig:log_mnist} shows a few of the most likely and unlikely (Fashion-)Mnist samples under the training data distribution. While samples are ordered by their marginal density $p(\x)$, the background light is proportional to their $\epsilon$-robustness, with darker colours for larger $\epsilon$. We can clearly see how these measures differ as, for example, although the model deems $1$s highly likely, $\epsilon$-robustness seems to vary with the shape/orientation of the trace.

Moreover, these two measures complement each other and allow us to better understand the underlying cause of the model's uncertainty. Notably, for a consistent model---one that fits the true data generating distribution if given sufficient data---and a sample $\x$ with high $p(\x)$ and low $\epsilon$-robustness, one may infer there is high aleatory uncertainty. 
A number 9 with an incomplete circle at the top is a good example of a pattern in handwritten digits that, albeit likely, is still hard to tell apart from a number 4. 
Conversely, an instance might be misshaped and hence unlikely, but still be associated to high robustness values. In that case, epistemic uncertainty is dominant, that is, the model has not been trained on similar examples and its high confidence estimate should not be trusted.
Distinguishing the two types of uncertainty is not only fundamental to better understand the task at hand but also to establish the correct course of action; namely, suspend judgement when faced with aleatory uncertainty or collect more data and possibly retrain the model in cases of epistemic uncertainty.

In all experiments\footnote{The source code will be available at the authors' web-page.}, the trees are made ``deep'', that is, we keep splitting the feature space until each leaf cell contains either only samples of one class or a single sample. That means the average depth of our models is $\Theta(\log n)$, with $n$ the number of samples in the training data \cite{Louppe2014}.
Such deep trees make \Gp{}s highly expressive, while the overall ensemble, by and large, avoids overfitting.
It is also worth noticing that our models are learned as regular Random Forests, with bootstrapping and gini-impurity criterion, and afterwards converted to \GeF{}s.
Moreover, we use fully-factorised leaves, $p(\X, Y)=p(X_1)\ldots p(X_m)p(Y)$, which are trivial to learn and compute but also achieve similar results as the original RF (identical predictions in each tree of the RF~\cite{Correia2020}).

\section{Conclusion}
While more experimentation is still needed, initial results indicate Generative Forests are a promising extension of Random Forests that effective leverages the properties of Probabilistic Circuits to detect out-of-domain samples and estimate the robustness of its own predictions.
We believe this is not only a important step towards more reliable machine learning but also a promising avenue for future research on deep hybrid discriminative-generative models.

\bibliography{GeFs.bib}
\bibliographystyle{icml2020}

\end{document}